\documentclass{article}

\usepackage[main, final]{neurips_2025}

\usepackage[utf8]{inputenc} 
\usepackage[T1]{fontenc}    
\usepackage{hyperref}       
\usepackage{url}            
\usepackage{booktabs}       
\usepackage{amsfonts}       
\usepackage{nicefrac}       
\usepackage{microtype}      
\usepackage[misc]{ifsym}
\usepackage{graphicx}
\usepackage[table]{xcolor}
\usepackage{amsmath}
\usepackage{cleveref}

\usepackage{algorithm}
\usepackage{algpseudocode}

\usepackage{xspace}
\usepackage{siunitx}
\usepackage{multirow}
\usepackage{subcaption}
\usepackage{bold-extra}

\usepackage{caption}

\DeclareCaptionLabelFormat{alglabel}{Algorithm} 
\makeatletter
\DeclareRobustCommand\onedot{\futurelet\@let@token\@onedot}
\def\@onedot{\ifx\@let@token.\else.\null\fi\xspace}
\def\eg{\emph{e.g}\onedot} 
\def\ie{\emph{i.e}\onedot}

\makeatother

\newsavebox\CBox
\def\textBF#1{\sbox\CBox{#1}\resizebox{\wd\CBox}{\ht\CBox}{\textbf{#1}}}
\newcommand{\bestscore}[1]{\textcolor{black}{\textBF{#1}}}

\newcommand{\Zhihang}[1]{\textcolor{black}{#1}}

\title{\Zhihang{Visionary}: The World Model Carrier Built on WebGPU-Powered Gaussian Splatting Platform}

\author{%
  Yuning Gong$^{1,2,}$\thanks{This work was initiated during an internship at the Shanghai AI Laboratory. {\scriptsize $^\textrm{\Letter}$} denotes the corresponding author (\texttt{zhongzhihang@pjlab.org.cn}).} \And
  Yifei Liu$^{1}$ \And
  Yifan Zhan$^{3}$ \And
  Muyao Niu$^{3}$ \And
  Xueying Li$^{4}$ \And
  Yuanjun Liao$^{1,2}$ \And
  Jiaming Chen$^{2}$ \And
  Yuanyuan Gao$^{1,5}$ \And
  Jiaqi Chen$^{1,5}$ \And
  Minming Chen$^{1}$ \And
  Li Zhou$^{1}$ \And
  Yuning Zhang$^{1}$ \And
  Wei Wang$^{1}$ \And
  Xiaoqing Hou$^{1}$ \And
  Huaxi Huang$^{1}$ \And
  Shixiang Tang$^{1}$ \And
  Le Ma$^{1}$ \And
  Dingwen Zhang$^{5}$ \And
  Xue Yang$^{4}$ \And
  Junchi Yan$^{1,4}$ \And
  Yanchi Zhang$^{2}$ \And
  Yinqiang Zheng$^{3}$ \And
  Xiao Sun$^{1}$ \And
  Zhihang Zhong$^{1,\,{\scriptsize \textrm{\Letter}}}$ \And
  \\
  $^{1}$Shanghai AI Laboratory \quad
  $^{2}$Sichuan University \quad
  $^{3}$The University of Tokyo \quad
  \\
  $^{4}$Shanghai Jiao Tong University
  $^{5}$Northwestern Polytechnical University\quad
  \\ \\
  Code: \href{https://github.com/Visionary-Laboratory/visionary}{https://github.com/Visionary-Laboratory/visionary}
}

\begin{document}

\maketitle

\begin{abstract}
Neural rendering, particularly 3D Gaussian Splatting (3DGS), has evolved rapidly and become a key component for building world models. However, existing viewer solutions remain fragmented, heavy, or constrained by legacy pipelines, resulting in high deployment friction and limited support for dynamic content and generative models. In this work, we present \textbf{Visionary}, an open, web-native platform for real-time various Gaussian Splatting and meshes rendering. Built on an efficient \texttt{WebGPU} renderer with per-frame \texttt{ONNX} inference, Visionary enables dynamic neural processing while maintaining a lightweight, ``click-to-run'' browser experience. It introduces a standardized Gaussian Generator contract, which not only supports standard 3DGS rendering but also allows plug-and-play algorithms to generate or update Gaussians each frame. Such inference also enables us to apply feedforward generative post-processing. The platform further offers a plug in \texttt{three.js} library with a concise TypeScript API for seamless integration into existing web applications. Experiments show that, under identical 3DGS assets, Visionary achieves superior rendering efficiency compared to current Web viewers due to GPU-based primitive sorting. It already supports multiple variants, including MLP-based 3DGS, 4DGS, neural avatars, and style transformation or enhancement networks. By unifying inference and rendering directly in the browser, Visionary significantly lowers the barrier to reproduction, comparison, and deployment of 3DGS-family methods, serving as a unified \textit{World Model Carrier} for both reconstructive and generative paradigms.
\end{abstract}
\section{Introduction}

\begin{figure}[!t]
  \centering
  \includegraphics[width= 1.0\linewidth ]{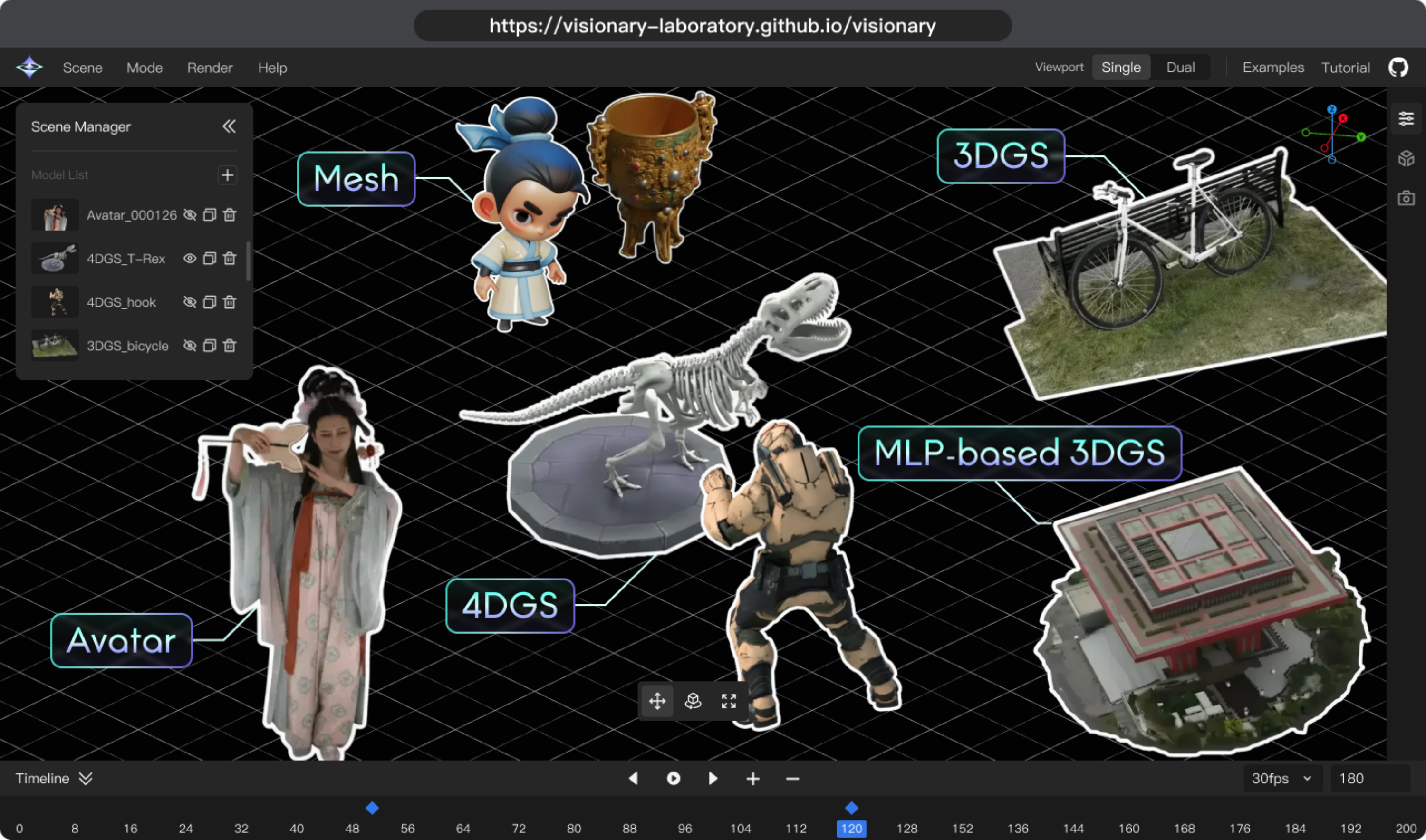}
  \caption{\textbf{Visionary as a Universal Runtime.} Visionary’s core system is packaged as a \texttt{three.js} plug-in for seamless extension and integration. As a demonstration, we present a lightweight web-based editor that runs directly in the browser: by simply visiting a URL, users can leverage local computing resources through \texttt{WebGPU} to efficiently and simultaneously render multiple heterogeneous 3D/4D Gaussian assets, while maintaining full compatibility with traditional mesh-based rendering pipelines.}
\end{figure}

Neural rendering, particularly 3D Gaussian Splatting (3DGS)~\cite{kerbl20233d}, has advanced rapidly in recent years and has become a key building block for world models~\cite{garrido2024learning}. Its efficiency and visual quality enable interactive, real-time 3D experiences that were previously impractical. However, a critical gap remains in deployment: while research focuses on training larger and more sophisticated models, the capability to distribute and execute these dynamic, and heterogeneous representations on consumer devices has lagged behind. Currently, there is an urgent need for a lightweight, flexible, and unified platform that can support the ever-expanding 3DGS ecosystem in a practical and accessible manner.

Existing systems suffer from significant deployment friction and limited extensibility. On the desktop, client-based frameworks such as SIBR~\cite{kerbl20233d} and engine-bound plug-ins for \href{https://unity.com/}{\texttt{Unity}}, \href{https://www.blender.org/}{\texttt{Blender}}, or \href{https://www.unrealengine.com/}{\texttt{Unreal Engine}} rely on heavy native stacks, including tightly coupled tool-chains and driver dependencies. While performant, these systems are difficult to configure, cumbersome to share, and poorly suited for rapid experimentation or integration of third-party algorithms. On the web, viewers such as \href{https://sparkjs.dev/}{\texttt{SparkJS}}, \href{https://superspl.at/editor}{\texttt{SuperSplat}}, and \href{https://github.com/mkkellogg/GaussianSplats3D}{\texttt{GaussianSplats3D}} are constrained by legacy \href{https://get.webgl.org/}{\texttt{WebGL}} pipelines, which restrict support for dynamic scenes, animatable avatars, and the integration of generative models. As a result, many demonstrations rely on precomputed Gaussians or server-side inference, reducing both interactivity and reproducibility.

Despite these limitations, web-based platforms remain highly attractive due to their accessibility and platform independence. This motivates a core challenge: how to design a browser-based system that can (i) support heterogeneous 3DGS algorithms without requiring users to re-engineer rendering pipelines or write low-level shader code, and (ii) tightly couple per-frame inference with high-throughput GPU rendering while maintaining real-time performance within a browser environment.

To address these challenges, we propose \textbf{Visionary}, a web-native \textit{World Model Carrier} built on a \href{https://github.com/gpuweb/gpuweb}{\texttt{WebGPU}}-based Gaussian Splatting platform. Visionary removes native dependency overhead by leveraging \texttt{WebGPU} as a unified compute and graphics backend, and standardizes algorithm integration through \href{https://onnx.ai/}{\texttt{ONNX}}-based per-frame inference. At its core is the Gaussian Generator contract, defined by a rigorous \texttt{ONNX} I/O schema and metadata, which specifies how algorithms generate or update Gaussian attributes, including position, scale, orientation, and color. Any 3DGS-family method can be loaded as a plug-in, executed directly in the browser, and rendered efficiently without modification to the underlying runtime.

Built on a fully browser-resident pipeline, Visionary unifies compute and rendering to support dynamic 3DGS workflows entirely on the client side. The current platform supports classic 3DGS, MLP-based 3DGS~\cite{lu2024scaffold}, neural avatars~\cite{hu2024gauhuman,zhan2025r3,qiu2025lhm}, and 4DGS~\cite{yang2024deformable} for dynamic scenes, along with generative post-processing networks~\cite{rombach2022high,wu2025difix3d+,chen2025exgs}. We further offer a \href{https://threejs.org/}{\texttt{three.js}} plug-in and exposed through a concise TypeScript API, enabling seamless integration with existing web applications. Across representative static and dynamic scenes, Visionary sustains real-time interactive performance and demonstrates substantial improvements in frame time, dynamic update latency, and scalability compared to existing web-based renderers.

The contributions of this work are as follows:
\begin{itemize}
\item Visionary, a web-native \textit{World Model Carrier} that unifies per-frame \texttt{ONNX} inference with a \texttt{WebGPU}-based Gaussian Splatting renderer and generative post-processing, enabling fully dynamic 3D neural rendering in the browser.
\item The Gaussian Generator contract, a standardized \texttt{ONNX} I/O and metadata interface that enables seamless plug-and-play integration of customized 3DGS algorithms without modifying the rendering pipeline.
\item A reference \texttt{WebGPU} implementation and \texttt{three.js} API, providing superior efficiency over \texttt{WebGL}-based viewers and straightforward integration into existing web engines for research and application development.
\end{itemize}
\section{Related Works}
\subsection{3D Gaussian Splatting}
Neural Radiance Fields (NeRF)~\cite{mildenhall2020nerf,barron2021mip,pumarola2021d,fridovich2022plenoxels,muller2022instant,li2023dynibar,zhan2024kfd} ignited the modern wave of neural rendering and corresponding down-stream applications~\cite{peng2021neural,weng2022humannerf,chen2024within}. However, volumetric NeRF rendering remains computationally expensive due to dense ray marching and repeated neural field evaluations. To address this inefficiency, 3D Gaussian Splatting~\cite{kerbl20233d} was introduced as an explicit, point-based alternative, which offer comparative render quality and extreme fast rendering speed. 

Building on the success of 3DGS, a rich ecosystem of extensions rapidly emerged. Methods such as 2DGS~\cite{zhang20242dgs}, PGSR~\cite{chen2024pgsr}, GOF~\cite{yu2024gaussian}, and RaDe-GS~\cite{zhang2024rade} provide more accurate depth rendering and establish tighter connections between Gaussian primitives and mesh-based representations. Structured 3D Gaussian like Scaffold-GS~\cite{lu2024scaffold} and Octree-GS~\cite{ren2024octree} further improve rendering quality and efficiency through structured anchors and neural decoding. While 3DGS is significantly more efficient than NeRF, scaling to city-level environments exposes new limitations in efficent training. Large-scale systems, including CityGaussian~\cite{liu2024citygaussian}, VastGaussian~\cite{lin2024vastgaussian}, Hier-GS~\cite{kerbl2024hierarchical}, and CityGS-X~\cite{gao2025citygs} propose specialized pipelines for training and rendering in expansive outdoor scenes. Concurrently, researchers have observed that Gaussian primitives often become overly fragmented and redundant, motivating a line of work on compression and pruning, such as LightGaussian~\cite{fan2024lightgaussian}, Compact-3DGS~\cite{lee2024compact}, and MaskGaussian~\cite{liu2025maskgaussian}. Recently, some studies~\cite{ye2025gaussian,gao2025proxy} have begun exploring the use of hardware rasterization techniques to accelerate the rendering process of 3DGS.
Beyond static scenes, dynamic extensions have emerged as well, including 4D Gaussian Splatting~\cite{yang2024deformable, wu20244d,bae2024per,cho20244d,huang2024sc,li2024spacetime,lu2024dn,shaw2024swings,yang2023real,chen2025alias}, and a series of work focusing on avatar reconstruction and animation~\cite{hu2024gauhuman,qian20243dgs,li2024animatable,niu2024bundle,qiu2025lhm, xu2025sequential,zhan2025towards,niu2025anicrafter}. More recently, a parallel thread of feedforward Gaussian reconstruction methods~\cite{charatan2024pixelsplat, chen2024mvsplat, li2024ggrt, jiang2025anysplat}, to bypass iterative optimization altogether and enable real-time reconstruction.

Overall, the 3DGS family has expanded rapidly, producing diverse representations, training paradigms, and rendering pipelines. This diversity highlights the need for a flexible, unified viewer architecture capable of visualizing, comparing, and mixing different Gaussian-based representations for research and content creation.

\subsection{3DGS Viewers and Plugins}
Effective visualization is crucial for the evaluation and deployment of neural rendering methods. 
Research-oriented frameworks typically employ desktop-based architectures to achieve high-performance rendering. The official SIBR viewer~\cite{kerbl20233d} relies on C++/CUDA to ensure efficiency but remains a standalone research tool. 
Similarly, Splatfacto, included in the \href{https://docs.nerf.studio/}{\texttt{Nerfstudio}}~\cite{tancik2023nerfstudio} framework, provides a browser-based interface for effective training monitoring. Nevertheless, as a client-server system coupled with the training engine, it necessitates a heavy local Python/CUDA backend and the original input data for initialization.
To leverage mature development ecosystems, plugins for game engines and content creation tools including \href{https://unity.com/}{\texttt{Unity}}, \href{https://www.unrealengine.com/}{\texttt{Unreal Engine}}, and \href{https://www.blender.org/}{\texttt{Blender}} were developed. These plugins allow 3DGS to interact with engine-specific features such as physics and lighting. 
However, both standalone viewers and engine-based plugins suffer from significant deployment friction. They typically require heavy installation stacks, specific GPU drivers (\eg, CUDA), and in the case of plugins, strict compatibility with specific engine versions. Such tight coupling and hardware dependency make these systems ill-suited for rapid sharing, lightweight reproduction, or cross-platform accessibility.

On the other hand, web-based viewers such as \href{https://github.com/mkkellogg/GaussianSplats3D}{\texttt{GaussianSplats3D}}, \href{https://superspl.at/editor}{\texttt{SuperSplat}}, and \href{https://sparkjs.dev/}{\texttt{SparkJS}} allow users to view static scenes directly in a browser without software installation. However, most existing web viewers rely on legacy \href{https://get.webgl.org/}{\texttt{WebGL}} pipelines. This dependence imposes severe performance bottlenecks, often necessitating CPU-based primitive sorting, which limits scalability for large scenes. Furthermore, the lack of flexible compute shader support in \texttt{WebGL} restricts these viewers to static assets, preventing the integration of dynamic Gaussian decoding or complex generative post-processing pipelines on the client side.

Compared with other desktop or web based solutions, our platform Visionary offers web-native accessibility and real-time \texttt{WebGPU} performance, while additionally supporting advanced 3DGS variants and generative models through a flexible \texttt{ONNX} interface.

\subsection{World Models and Interactive Generative Video}
A world model aims to internalize the governing laws of a physical environment, learning to predict future states based on current observations and actions.
This capability holds transformative potential spanning a wide range of verticals, promising to enable high-fidelity industrial digital twins, immersive entertainment and even personalized interactive education.
To facilitate scalability, the dominant paradigm of world models bypasses explicit 3D reconstruction, instead learning auto-regressively from vast video data to simulate the ``world'' as a sequence of 2D latent frames, manifested as Interactive Generative Video (IGV).
Prominent general-purpose foundation models, such as \href{https://deepmind.google/blog/genie-3-a-new-frontier-for-world-models/}{Genie 3}, V-JEPA 2~\cite{assran2025v}, and Cosmos 2.5~\cite{ali2025world}, exemplify this approach, demonstrating remarkable capabilities in synthesizing diverse visual dynamics.
This methodology has seen rapid adoption across specific domains as well, ranging from autonomous driving~\cite{wang2024drivedreamer,gao2023magicdrive,li2024drivingdiffusion,wen2024panacea,zhao2025drivedreamer,gao2024vista,hassan2025gem,zhang2025epona,russell2025gaia,ren2025cosmos} and embodied robotic manipulation~\cite{ajay2022conditional,zhou2024robodreamer,shang2025roboscape,lyu2025dywa,zheng2025flare} to open-world gaming agents~\cite{xiao2025worldmem,decart2024oasis,guo2025mineworld,alonso2024diffusion} and science~\cite{zhang2025cellflux,yang2025xray2xray,yang2025twinmarket}.
However, lacking explicit spatial understanding, these models suffer from inherent 3D inconsistency, frequently exhibiting ``geometric hallucinations'' and failing to maintain consistent object identity.
Consequently, the generated ``worlds'' often lack the physical realism and long-horizon stability essential for rigorous simulation.

Leveraging advancements in feed-forward Gaussian splatting~\cite{charatan2024pixelsplat,chen2024mvsplat,jiang2025anysplat} and VGGT~\cite{wang2025vggt}, recent research~\cite{ren2025gen3c,ma2025you,wu2025video,li2025vmem} integrates 3D priors into the latent state space to enforce physical plausibility.
Unlike transient 2D features, explicit 3D intermediate states ensure superior multi-view consistency, allowing the ``camera'' to move freely without breaking the scene's illusion while providing a robust global memory bank.
However, realizing the full potential of such high-fidelity representations requires accessible visualization tooling.
Our platform addresses this infrastructural need by offering a standardized \texttt{ONNX}-based pipeline for the real-time inspection of explicit 3D states.
Furthermore, its inherent compatibility with IGV workflows makes it an ideal testbed for developing next-generation, physics-aware World Models.
\section{Pipeline}

\begin{figure}[!t]
  \centering
  \includegraphics[width= 1.0\linewidth ]{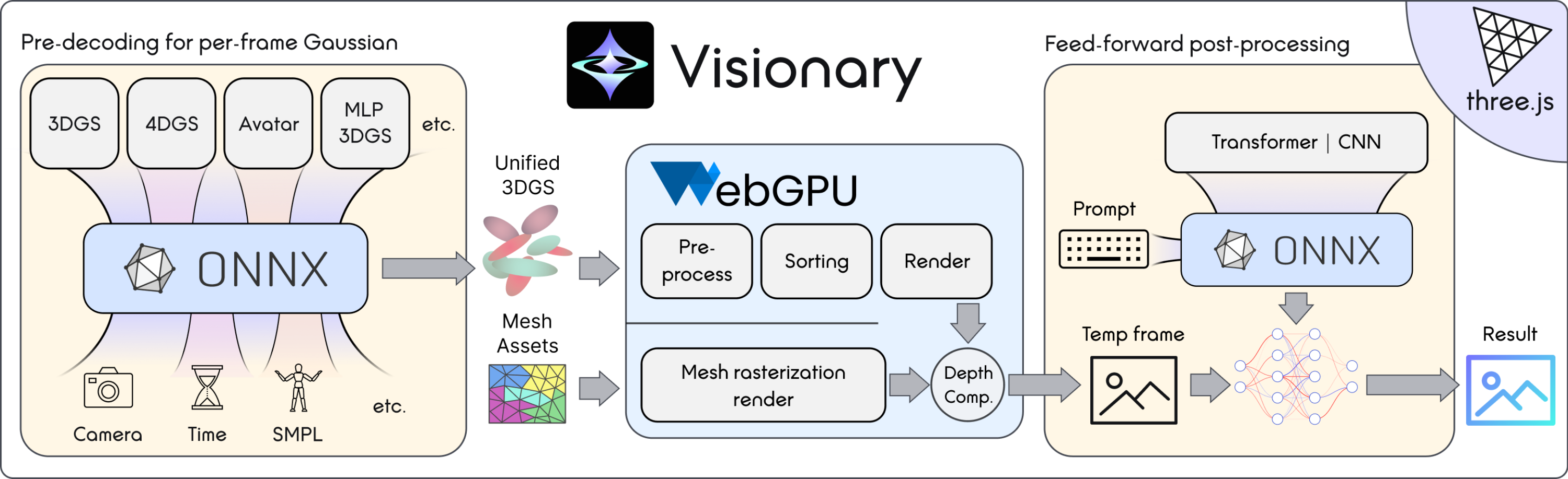}
  \caption{\textbf{Pipeline of Visionary.} Visionary first performs pre-decoding through \texttt{ONNX} to support diverse 3DGS variants. The generated Gaussians are then combined with optional mesh assets and passed into a \texttt{WebGPU}-based hybrid rendering pipeline, where depth-aware composition produces the final image. In the post-processing stage, \texttt{ONNX}-based generative models can be optionally applied for tasks such as stylization and enhancement, yielding the final output.}
  \label{fig:pipeline}
\end{figure}

\subsection{Overview}
\Zhihang{
The overall design of Visionary follows three guiding principles. 
\textbf{Web-native first}: all components run directly in the browser via \texttt{WebGPU}, enabling zero-install sharing, on-device inference for privacy, and consistent behavior across operating systems. 
\textbf{Contract-driven extensibility}: algorithms are exported to \texttt{ONNX}, while the runtime handles scheduling, memory management, and rendering, eliminating the need for per-algorithm code branches. 
\textbf{Real-time coupling}: inference and rendering are executed within a single frame budget, enabling fast interaction. 
The overall pipeline is illustrated in \Cref{fig:pipeline}. 
We first review the definition of 3DGS in \Cref{sec:3dgs_definition}. 
In \Cref{sec:predecoding}, we present the Gaussian Generator contract built on \texttt{ONNX} and demonstrate several representative 3DGS variants.
We then describe the \texttt{WebGPU}-based rendering pipeline for hybrid 3DGS and mesh visualization in \Cref{sec:webgpu_renderer}. 
Finally, in \Cref{sec:postprocessing}, we introduce how generative models can also be integrated through \texttt{ONNX} for post-processing.
}

\subsection{Definition of 3DGS}
\label{sec:3dgs_definition}
Gaussian Splatting represents a scene as a set of $N$ anisotropic 3D Gaussian primitives:
$
\mathcal{G} = \{ G_i \}_{i=1}^{N},
$
where each Gaussian $G_i$ is defined by a mean position $\boldsymbol\mu_i \in \mathbb{R}^3$, a covariance matrix $\boldsymbol{\Sigma}_i \in \mathbb{R}^{3 \times 3}$, which is derived from a scale $\boldsymbol s_i \in \mathbb{R}^3$ and a rotation quaternion $\boldsymbol r_i \in \mathbb{R}^4$, an opacity $\alpha_i \in (0,1)$, and a view‐independent color $\mathbf{c}_i \in [0,1]^3$ encoded by spherical harmonics.

For a camera with projection operator $\Pi(\cdot)$, each 3D Gaussian is projected to a 2D elliptical Gaussian
\begin{equation}
\mathbf{x}_i = \Pi(\boldsymbol\mu_i), \qquad
\mathbf{S}_i = \mathbf{J}_i \boldsymbol{\Sigma}_i \mathbf{J}_i^\top,
\end{equation}
where $\mathbf{J}_i$ denotes the Jacobian of $\Pi$ at $\boldsymbol{\mu}_i$.
Given a pixel location $\mathbf{x}$, its contribution from Gaussian $i$ is modeled as
\begin{equation}
w_i(\mathbf{x}) = \alpha_i \exp\!\left(
-\frac{1}{2} (\mathbf{x} - \mathbf{x}_i)^\top
\mathbf{S}_i^{-1}
(\mathbf{x} - \mathbf{x}_i)
\right).
\end{equation}

Following front‐to‐back alpha compositing, the rendered color at pixel $\mathbf{x}$ is computed as
\begin{equation}
\mathbf{C}(\mathbf{x}) = 
\sum_{i=1}^{N}
\left(
w_i(\mathbf{x})
\prod_{j < i} (1 - w_j(\mathbf{x}))
\right)
\mathbf{c}_i,
\end{equation}
where Gaussians are sorted by depth w.r.t.\ the camera.
This splatting formulation enables differentiable rendering and efficient optimization of all Gaussian parameters through gradient descent.

\subsection{\texttt{ONNX}-based Gaussian Pre-decoding}
\label{sec:predecoding}
3DGS has many variants, each adopting different algorithms to produce the final rendering. For example, MLP-based 3DGS~\cite{lu2024scaffold} decodes Gaussians from anchors according to a give camera view and then rasterizes the decoded Gaussians, while 4DGS~\cite{wu20244d} computes a deformation field based on a timestamp and then rasterizes the deformed Gaussians. To avoid modifying the renderer for each existing or newly proposed method, we deciced to split the rendering logic into two parts: a method-specific pre-decoding stage and a unified renderer.

To support heterogeneous 3DGS-family methods without modifying the renderer, Visionary adopts an \texttt{ONNX}-based \emph{pre-decoding} stage, where each algorithm is exported as a plugin that generates or updates Gaussian attributes every frame. 
We formalize this interface as the \textbf{Gaussian Generator} contract, specified by a fixed set of \texttt{ONNX} I/O tensors and accompanying metadata, enabling contract-driven extensibility. ONNX (Open Neural Network Exchange) serves as an open standard for machine learning interoperability, defining a common computation graph representation independent of specific frameworks. By leveraging this ecosystem, models trained in diverse environments (\eg., PyTorch or TensorFlow) can be unified and deployed seamlessly across different backends.

\paragraph{Contract I/O.}
At runtime, the client provides lightweight per-frame inputs (\eg, frame index, control signals or camera). The \texttt{ONNX} graph outputs a variable number of Gaussians $\mathcal{G}_t$ in a packed layout (position, opacity, upper-covariance, and appearance), along with model-level metadata such as number of points and data types (\texttt{FP32}/\texttt{FP16}). The renderer consumes these outputs directly as storage buffers for the \texttt{WebGPU} pipeline.

\paragraph{Why pre-decoding?}
By encapsulating method-specific neural decoding and deformation logic inside \texttt{ONNX}, Visionary avoids per-method shader branches and heavy native dependencies. For example, in avatar animation, the skeletal forward kinematics and per-Gaussian deformation can be exported into the \texttt{ONNX} graph, so the browser only feeds compact pose parameters while receiving deformed Gaussians ready for rasterization. 

\paragraph{Representative variants.}
In the following subsections, we instantiate this contract for multiple families, including MLP-based
3DGS, 4DGS, and neural avatars, demonstrating that diverse representations can be unified under the
same pre-decoding $\rightarrow$ rendering pipeline.

\paragraph{Deployment-oriented graph optimizations.}
Beyond unifying heterogeneous variants under a single \texttt{ONNX} pre-decoding contract, we introduce two
practical optimizations to improve real-time robustness on \texttt{ONNX} Runtime \texttt{WebGPU}.

\textbf{(i) Enabling \texttt{capture\_graph}.}
For per-frame generators with repetitive execution patterns, we revise the runtime scheduling and I/O
binding logic to make the inference path compatible with \texttt{WebGPU} graph capture. Concretely, we keep the session and bindings stable across frames (with an optional warm-up run) so that the runtime can
reuse the captured execution graph, reducing JavaScript-side dispatch overhead and stabilizing frame time.

\textbf{(ii) Post-export rewriting for large \texttt{Concat}/\texttt{Split} patterns.}
In some exported models, a large number of per-slot tensors are concatenated or split in a single
operator (\eg, one-shot \texttt{Concat} over many slots, followed by \texttt{Split} for downstream
consumers). Such patterns can be fragile under current \texttt{WebGPU} limits. We therefore apply a lightweight
post-processing pass on the exported \texttt{ONNX} graph to partition these operations into smaller, equivalent
chunks (\eg, chunked concat/slice-based splits), producing a \texttt{WebGPU}-friendly graph while preserving the same packed output layout required by the Gaussian Generator contract.

\subsubsection{Support for MLP-based 3DGS}
MLP-based 3DGS like Scaffold-GS~\cite{lu2024scaffold} and Octree-GS~\cite{ren2024octree} have been proposed to use MLP to decode Gaussians from anchors that store shared features of nearby Gaussians, which offers better visual quality and reduced storage. We leverage Scaffold-GS~\cite{lu2024scaffold} to demonstrate the methodology for exporting such representation into the \texttt{ONNX} format and integrating into our platform. Instead of directly reconstructing Gaussians from sparse SfM points, these methods first extract a sparse voxel grid and place anchors at the centers of occupied voxels. Each anchor is associated with a feature vector $f_i$, which is fed into a multi-layer perceptron (MLP) to generate neural Gaussian parameters:
\begin{equation}
\left\{ (\mu_j, \Sigma_j, c_j, \alpha_j) \mid j \in \mathcal{M} \right\} = \left\{ \mathrm{MLP}_\theta(f_i, d_{\mathrm{view}}) \mid i \in \mathcal{N} \right\},
\label{eq:anchor_mlp}
\end{equation}
where $\theta$ denotes the learnable MLP weights, and $\mu_j$, $\Sigma_j$, $c_j$, and $\alpha_j$ represent the mean, covariance, color, and opacity of the $j$-th neural Gaussian generated from the $i$-th anchor under viewing direction $d_{\mathrm{view}}$. These neural Gaussians are then rasterized using the standard 3DGS formulation.

While MLP-based 3DGS methods exhibit strong rendering flexibility, Eq.~\Cref{eq:anchor_mlp} reveals a key challenge: the neural Gaussian parameters must be re-generated for every frame, making real-time online visualization difficult. To the best of our knowledge, no existing system provides an interactive, in-browser viewer for such dynamically generated Gaussians.

To address this limitation, we integrate the MLP-based 3DGS pipeline with \texttt{ONNX} Runtime \texttt{WebGPU} and enable efficient, browser-side visualization. Specifically, we export the trained MLP, along with the anchor positions, scales, and feature vectors, into a static \texttt{ONNX} compute graph. This design allows the viewer to compute neural Gaussian parameters on-the-fly for arbitrary camera poses, enabling fast, real-time visualization directly in the browser.

\subsubsection{Support for 4DGS}

To address the challenge of reconstructing and rendering dynamic scenes, mainstream \href{https://github.com/hustvl/4DGaussians?tab=readme-ov-file}{4D Gaussians} approaches typically circumvent the prohibitive storage overhead of storing Gaussian parameters for every individual frame.
Instead, their modeling often combines a \textbf{canonical space} with a \textbf{neural deformable field}.
We leverage 4D Gaussians~\cite{wu20244d} as a representative case to demonstrate the methodology for exporting such deformable fields into the \texttt{ONNX} format.

Specifically, 4D Gaussians employs an efficient HexPlane representation (akin to K-Planes~\cite{kplanes_2023}) for spatiotemporal feature encoding. Rather than relying on computationally intensive implicit MLP to model the deformation field, this method factorizes the 4D manifold into six orthogonal 2D feature planes (\ie, $P_{xy}, P_{xz}, P_{yz}, P_{xt}, P_{yt}, P_{zt}$).
For each Gaussian primitive in the canonical space, its coordinates and current timestamp $(x, y, z, t)$ are projected onto these multi-resolution planes to retrieve corresponding feature vectors via bilinear interpolation.
These features are subsequently concatenated and fed into a lightweight multi-head MLP decoder $\varphi$ to predict the deformation attributes, specifically position offsets $(\Delta x, \Delta y, \Delta z)$, rotation corrections $\Delta r$, and scaling variations $\Delta s$. This process effectively transforms the static canonical Gaussians into their time-dependent states for real-time rendering.

To integrate {4D Gaussians}~\cite{wu20244d} into our platform, we established a comprehensive training-to-export pipeline.
We developed a custom wrapper that encapsulates the deformation network, comprising multi-resolution feature grids $P$ and the MLP decoder $\varphi$, into a standalone \texttt{torch.nn.Module}.
This design effectively decouples the inference logic from training-specific rasterization operators, ensuring compatibility with standard \texttt{ONNX} runtimes.
Notably, our exported model embeds the canonical Gaussian attributes directly as internal constant tensors (Initializers), rather than requiring external inputs.
Consequently, the model accepts a scalar timestamp $t$ as the sole input to query the HexPlane and output deformation residuals.
This self-contained architecture obviates the need for external geometry files (\eg, PLY), streamlining the web client's reconstruction workflow via a unified \texttt{ONNX} model.

\subsubsection{Support for Animatable Human Avatar}

Our platform supports a broad class of animatable human avatar methods that utilize a \textbf{canonical space} formulation decoupled from the deformation logic.
In this paradigm, the scene is represented by a set of static canonical 3D Gaussians in a neutral pose (\eg., T-pose), which are dynamically warped to the observation space driven by a parametric body model (\eg, SMPL-X~\cite{SMPLX2019}).

Let the canonical representation be denoted as $\mathcal{G}_{can} = \{ \hat{G}_i \}_{i=1}^N$, where each primitive $\hat{G}_i$ typically contains a mean position $\hat{\boldsymbol{\mu}}_i$, a covariance $\hat{\boldsymbol{\Sigma}}_i$, and learnable skinning weights $\mathcal{W}_i = \{w_{i,k}\}_{k=1}^{K}$ associated with $K$ skeletal joints.
The deformation is generally modeled via Linear Blend Skinning (LBS). Given the body pose parameters $\boldsymbol{\theta}$ and shape parameters $\boldsymbol{\beta}$ for a specific frame, the parametric model defines a global transformation matrix $\mathbf{M}_k(\boldsymbol{\theta}, \boldsymbol{\beta}) \in SE(3)$ for each joint $k$.
The local transformation $\mathbf{T}_i$ for the $i$-th Gaussian is computed by aggregating these joint transformations:
\begin{equation}
\mathbf{T}_i(\boldsymbol{\theta}, \boldsymbol{\beta}) = \sum_{k=1}^{K} w_{i,k} \mathbf{M}_k(\boldsymbol{\theta}, \boldsymbol{\beta}) =
\begin{bmatrix}
\mathbf{R}_i & \mathbf{t}_i \\
\mathbf{0} & 1
\end{bmatrix},
\end{equation}
where $\mathbf{R}_i$ and $\mathbf{t}_i$ represent the blended rotation and translation, respectively.
Consequently, the deformed attributes for the live frame are updated as:
\begin{equation}
\begin{aligned}
\boldsymbol{\mu}_i &= \mathbf{T}_i \cdot \hat{\boldsymbol{\mu}}_i, \\
\boldsymbol{\Sigma}_i &= \mathbf{R}_i \hat{\boldsymbol{\Sigma}}_i \mathbf{R}_i^\top.
\end{aligned}
\end{equation}
This formulation allows the renderer to handle diverse avatar methods that adhere to the canonical-deformation schema.

As a representative implementation of this paradigm, we integrate \href{https://github.com/aigc3d/LHM}{LHM}~\cite{qiu2025lhm} into our \texttt{ONNX}-based pipeline. LHM learns a high-fidelity canonical representation driven by SMPL-X.
To enable web-based rendering, we export the deformation logic into a specialized \texttt{ONNX} graph.
In this setup, the heavy skeletal computations including the forward kinematics of the SMPL-X skeleton and the per-Gaussian LBS operations defined above are encapsulated within the model.
The canonical Gaussian attributes ($\hat{\boldsymbol{\mu}}, \hat{\boldsymbol{\Sigma}}$) and skinning weights are baked into the model as initializers (constants).
During runtime, the client simply feeds the lightweight SMPL-X parameters ($\boldsymbol{\theta}, \boldsymbol{\beta}$) into the \texttt{ONNX} runtime. The model then acts as a pre-decoder, outputting the transformed positions and covariances directly to the rasterizer, thereby achieving real-time animation without heavy client-side dependency management.

We further adapt \href{https://github.com/Yifever20002/R3Avatars}{R3Avatar}~\cite{zhan2025r3} to support high-fidelity novel view rendering of human avatars.
In this configuration, we likewise pre-calculate and encapsulate the complex skeletal dynamics and neural deformation logic within the \texttt{ONNX} graph.
The client provides a frame index as input.
Internally, the \texttt{ONNX} runtime maps this index to the corresponding temporal state, decodes the specific avatar configuration, and outputs the deformed Gaussians directly to the rasterizer.
\algrenewcommand\algorithmicrequire{\textbf{Require:}}
\algrenewcommand\algorithmicensure{\textbf{Ensure:}}

\begin{algorithm}[t]
\caption{Pseudo code for \texttt{WebGPU} 3DGS rendering}
\label{alg:webgpu-3dgs}
\begin{algorithmic}
\Require Gaussians Models: $\mathcal{M}_k = \{(\mu_i,s_i,r_i,c_i)\}_{i=1}^{N_k}$; camera $(V,P)$; per-model transform $M_k$
\Ensure Rendered frame

\State \textbf{/* Pre-pack (only once) */} 
\ForAll{gaussian $i$}
  \State $\Sigma_i \gets \mathrm{CovFromScaleRot}(s_i,r_i)$
  \State store $\mathrm{upper}(\Sigma_i)$, $\mu_i$ and $c_i$ as \texttt{FP16}-packed \texttt{u32} buffers
\EndFor

\State \textbf{/* Per-frame */} \Comment{global buffers + atomic count}
\State $\texttt{count}\gets 0$; $\texttt{dispatch}\gets 0$ \Comment{reset atomics/indirects}

\ForAll{model $\mathcal{M}_k$}
  \State bind \texttt{ModelParams}($M_k$, offsets, types, SHdeg, \dots)
  \State dispatch \Call{PreprocessKernel}{$\mathcal{M}_k$}
\EndFor
\State \Call{Radix Sort}{\texttt{depths}, \texttt{indices}, \texttt{count}}
\State \Call{RenderMeshDepthPrepass}{} \Comment{rasterize mesh, output depth texture $D_\text{mesh}$}
\State \Call{DrawSplatsWithMeshDepth}{\texttt{points2d}, \texttt{indices}, \texttt{count}, $D_\text{mesh}$}
\Comment{premultiplied $\alpha$, back-to-front}

\Procedure{Preprocess}{idx}
  \State $(p,\alpha)\gets$ read pos/opacity; $\Sigma^{(6)}\gets$ read upper-cov
  \State $x_w \gets M_k\,[\mu;1]$; $x_c\gets V\,x_w$; $x_{clip}\gets P\,x_c$
  \If{frustum-cull fails \textbf{or} $\alpha<\alpha_{\min}$} \State \Return \EndIf
  \State compute 2D ellipse $(v_1,v_2)$ and NDC center $(u,v,z)$
  \State $rgba \gets$ evaluate color (raw/SH) \Comment{pack to \texttt{FP16}}
  \State $k \gets \Call{atomicAdd}{\texttt{count}, 1}$
  \State write \texttt{points2d}[$k$], \texttt{depths}[$k$], \texttt{indices}[$k$]
\EndProcedure
\end{algorithmic}
\end{algorithm}
\subsection{\texttt{WebGPU}-based Hybrid Renderer}
\label{sec:webgpu_renderer}

    Visionary renders multiple 3DGS models together with optional mesh assets in a single \texttt{WebGPU} pipeline, and composes them via depth-aware composition. 
Each Gaussian primitive stores position $\boldsymbol\mu_i$, scale $\boldsymbol s_i$, rotation $\boldsymbol r_i$, and appearance $\boldsymbol c_i$ (raw RGB or SH).
For efficiency, we pre-pack Gaussian attributes into GPU-friendly buffers and execute per-frame screen-space preprocessing and sorting entirely on the GPU.

\paragraph{Pre-pack and \texttt{FP16} layout.}
Given a Gaussian parameter unit $(\mu_i,s_i,r_i,c_i)$, $s_i$ and $r_i$ are converted into a symmetric covariance $\Sigma_i \in \mathbb{R}^{3\times 3}$,
and the upper-triangular 6-tuple are stored as $\Sigma_i^{(6)}$.
We then cast $(\mu_i,\alpha_i,\Sigma_i^{(6)},c_i)$ to \texttt{fp16} and pack every two units into \texttt{u32} storage to reduce bandwidth.

\paragraph{Per-frame preprocessing.}
For each loaded model $\mathcal{M}_k = \{(\mu_i,s_i,r_i,c_i)\}_{i=1}^{N_k}$, we dispatch a compute kernel that (i) applies a user-defined affine transformation
$M_k$ to $\mathcal{M}_k$, (ii) transforms its parameter units to camera space and clip space, (iii) performs frustum and opacity culling to remove invalid splats, and
(iv) computes the 2D ellipse eigenvectors $(v_1,v_2)$ and NDC center $(u,v,z)$ (written as \texttt{Splat}).
Valid splats are appended into global buffers via an atomic counter and simultaneously write depth values, \ie the $z$ in NDC space, as keys for sorting in the next step.

\paragraph{GPU sorting and rasterization.}
After preprocessing, we perform a GPU radix sort~\cite{mcilroy1993engineering} over depths and then draw all splats using
instanced rasterization. The vertex shader expands each splat into a screen-space quad using $(v_1,v_2)$,
while the fragment shader evaluates the Gaussian weight and outputs pre-multiplied color.

\paragraph{Depth-aware composition with mesh.}
If a mesh is present, we first rasterize the mesh to obtain a depth buffer $D_{\text{mesh}}$ (and optionally
its color). During Gaussian rasterization, we keep \emph{depth test enabled but depth write disabled}:
Gaussian fragments with $z_{\text{gs}} > D_{\text{mesh}}$ are rejected (occluded by mesh), while visible fragments
are alpha-composited in back-to-front order.
\subsection{ONNX-based Post-processing}
\label{sec:postprocessing}

In our \texttt{ONNX-based}-based post-processing workflow, we export the diffusion denoiser (U-Net~\cite{ronneberger2015u,ho2020denoising}) to a Web-friendly \texttt{ONNX} graph. Here, we take EXGS~\cite{chen2025exgs} as an example of post processing enhancement and talk about how to export it. Concretely, we load the model and wrap it with a lightweight adapter that standardizes timestep handling, either using a fixed timestep buffer for a specific denoising step or exposing the timestep as a dynamic input. The model is then switched to the evaluation mode and optionally cast to \texttt{FP16} on the target device for efficient inference.

We next construct shape-correct dummy inputs and optionally enable dynamic axes. The exported \texttt{ONNX} can then run with varying inference shapes. After a sanity-check forward pass, we invoke \texttt{torch.onnx.export()} with appropriate input/output names, opset version, and constant folding. The exported model is finally validated via \texttt{onnx.checker.check\_model()}.

\section{Experiments}
\label{sec:experiments}

\begin{figure}[!t]
    \centering
    \begin{subfigure}[b]{0.32\textwidth}
        \centering
        \includegraphics[width=\textwidth]{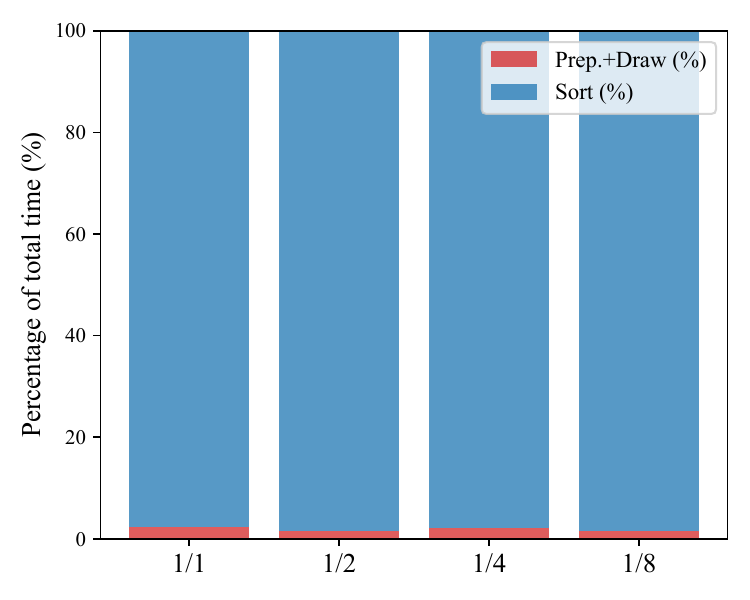}
        \caption{\texttt{SparkJS} time breakdown.}
        \label{fig:spark_percent}
    \end{subfigure}
    \hfill
    \begin{subfigure}[b]{0.32\textwidth}
        \centering
        \includegraphics[width=\textwidth]{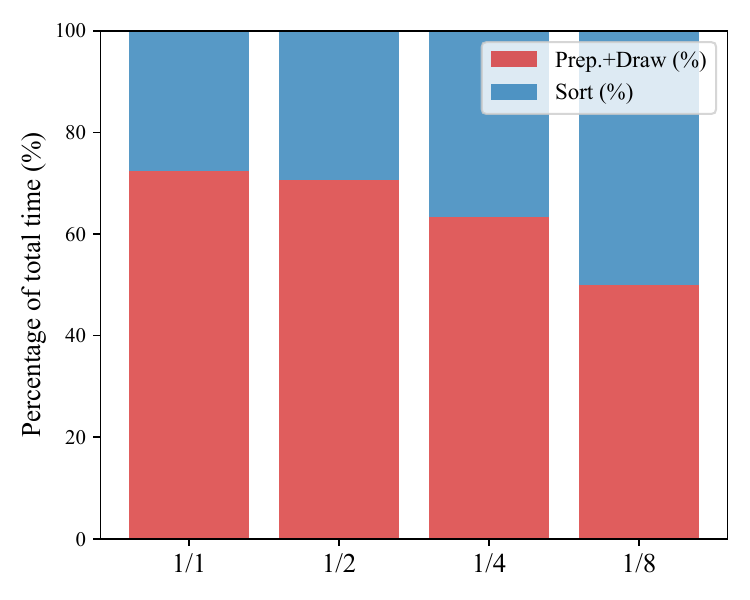}
        \caption{Visionary time breakdown.}
        \label{fig:ours_percent}
    \end{subfigure}
    \hfill
    \begin{subfigure}[b]{0.32\textwidth}
        \centering
        \includegraphics[width=\textwidth]{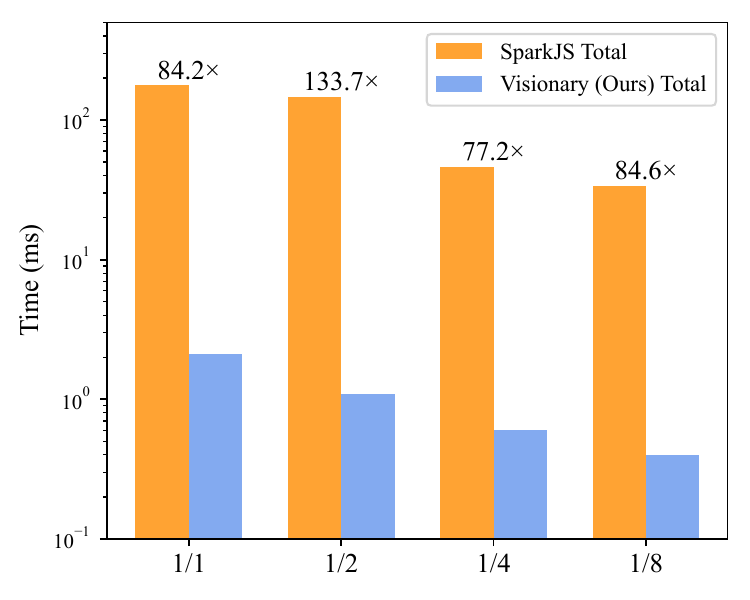}
        \caption{Total time (log scale).}
        \label{fig:total_compare}
    \end{subfigure}
    \caption{\textbf{Runtime comparison.} Experiments are done under identical Gaussian complexity using the classic ``bicycle'' scene of 3DGS~\cite{kerbl20233d} (6\si{M} Gaussians at full resolution, and $1/2$, $1/4$, $1/8$ scales). 
    (a): \texttt{SparkJS} shows a dominant CPU sorting cost.
    (b): Visionary shifts computation to GPU with low and stable overhead.
    (c): Log-scale comparison shows up to $\sim$100$\times$ speed-up over \texttt{SparkJS}.
    }
    \label{fig:runtime_comparison}
\end{figure}

\subsection{Setup}
We evaluate Visionary as a web-native renderer that couples per-frame Gaussian generation (\texttt{ONNX}) with high-throughput \texttt{WebGPU} splatting. We focus our evaluation on two aspects:
(i) end-to-end rendering efficiency and scalability on static 3DGS assets, and
(ii) visual robustness of alpha-composited splatting under challenging interaction patterns (\eg, rapid view changes)
and multi-component composition.
We compare against representative \texttt{WebGL}-based viewers, including \texttt{SparkJS} and \texttt{SuperSplat}.
Unless otherwise stated, all comparisons are conducted using identical 3DGS assets and camera trajectories. Also, we report the runtime of different variants implemented with our \texttt{WebGPU} renderer.
All experiments are conducted on a workstation equipped with an NVIDIA RTX 4090 GPU and an Intel w5-3435X CPU.

\begin{table}[!t]
\centering
\setlength{\tabcolsep}{3pt}
\caption{\textbf{Time analysis.} The sorting process in \texttt{SparkJS} is performed on the CPU. The GPU execution time is obtained using PIX replay, where we first capture a representative frame in PIX and then replay it within PIX to profile performance.}
\label{tab:time_analysis}
\begin{tabular}{lcccccc}
\toprule
\multirow{2}{*}{\# GS [\si{M}]} & \multicolumn{3}{c}{\href{https://sparkjs.dev/}{\texttt{SparkJS}}} & \multicolumn{3}{c}{Visionary (Ours)} \\
\cmidrule(lr){2-4}
\cmidrule(lr){5-7}
& Sort [\si{ms}] & Prep.+Draw [\si{ms}] & Total [\si{ms}] & Sort [\si{ms}] & Prep.+Draw [\si{ms}] & Total [\si{ms}] \\
6.062 (1/1) & 172.87 & 4.03 & 176.90 & \bestscore{0.58} & \bestscore{1.52} & \textbf{2.09} \\ 
3.031 (1/2) & 143.50 & 2.25 & 145.75 & \bestscore{0.32} & \bestscore{0.77} & \textbf{1.09} \\
1.515 (1/4) & 45.27 & 1.03 & 46.29 & \bestscore{0.22} & \bestscore{0.38} & \textbf{0.60} \\
0.758 (1/8) & 33.31 & 0.52 & 33.82 & \bestscore{0.20} & \bestscore{0.20} & \textbf{0.40} \\
\bottomrule
\end{tabular}
\end{table}

\begin{table}[!th]
\centering
\setlength{\tabcolsep}{9pt}
\caption{\textbf{Rendering quality analysis.} Evaluated on MipNeRF360~\cite{barron2022mip} dataset.}
\label{tab:render_quality}
\begin{tabular}{lccc}
\toprule
& PSNR $\uparrow$ & SSIM $\uparrow$ & LPIPS $\downarrow$ \\ 
\midrule
\href{https://sparkjs.dev/}{\texttt{SparkJS}} & 27.315 & 0.825 & 0.253 \\
Visionary (Ours) & \bestscore{27.867} & \bestscore{0.828} & \bestscore{0.249} \\
\bottomrule
\end{tabular}
\end{table}

\subsection{Runtime and Scalability vs.\ \texttt{SparkJS}}
\label{sec:exp_runtime_spark}
We first benchmark a standard large-scale static scene: the classic \texttt{bicycle} 3DGS asset
with $\sim$6\si{M} Gaussians at full resolution, and its downscaled variants (1/2, 1/4, 1/8).
\Cref{fig:runtime_comparison} reports the frame-time breakdown and the end-to-end latency in \Cref{tab:time_analysis}. To notice, we measure the GPU execution time using PIX replay, where we first capture a representative frame in PIX and then replay it within PIX to profile performance. 

\texttt{SparkJS} is bottlenecked by CPU-side primitive, which dominates the total frame time.
In contrast, Visionary moves per-frame preprocessing and global sorting to \texttt{WebGPU} compute,
resulting in low and stable overhead across resolutions. Concretely, at full resolution (6.062\si{M} Gaussians),
\texttt{SparkJS} spends 172.87\si{ms} on sorting and 176.90\si{ms} total per frame, while Visionary reduces this to
0.58\si{ms} sorting and 2.09\si{ms} total per frame. Across resolutions, Visionary achieves up to $\sim$135$\times$
end-to-end speedup over \texttt{SparkJS} (\Cref{fig:runtime_comparison}).

\subsection{Rendering Quality}
Speed improvements should not come at the cost of image quality.
On the MipNeRF360~\cite{barron2022mip} benchmark, Visionary matches (and slightly improves) the rendering quality compared to \texttt{SparkJS} in termps of PSNR, SSIM and LPIPS as in \cref{tab:render_quality}. This slight improvement can be attributed to two design choices: Visionary avoids the aggressive quantization used in \texttt{SparkJS} to preserve more detail while still maintaining interactive speed, and our \texttt{WebGPU} implementation performs 3DGS preprocessing in compute shaders rather than relying on a rasterization-based \texttt{WebGL} pipeline.
 
\begin{figure}[!th]
  \centering
  \includegraphics[width=\linewidth]{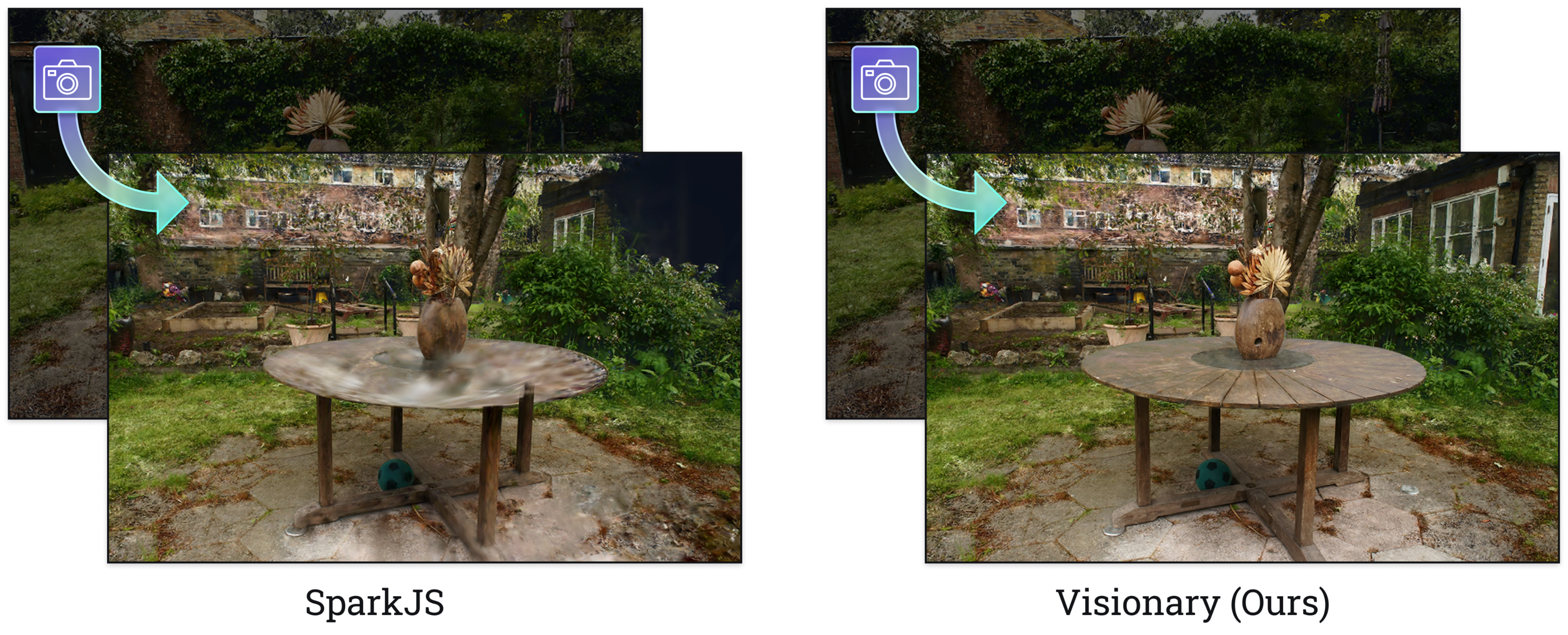}
  \caption{\textbf{Artifacts caused by lazy sorting in \texttt{SparkJS}.}
  When rotating the viewpoint rapidly, the stale/incrementally updated order can become invalid,
  producing incorrect alpha compositing and visible temporal artifacts. The efficient implementation of Visioanry avoids this flaw. The version of \texttt{SparkJS} tested here is the latest available, \texttt{v0.1.10}.}
\label{fig:spark_lazy_sort_artifacts}
\end{figure}

\begin{figure}[!th]
  \centering
  \includegraphics[width=\linewidth]{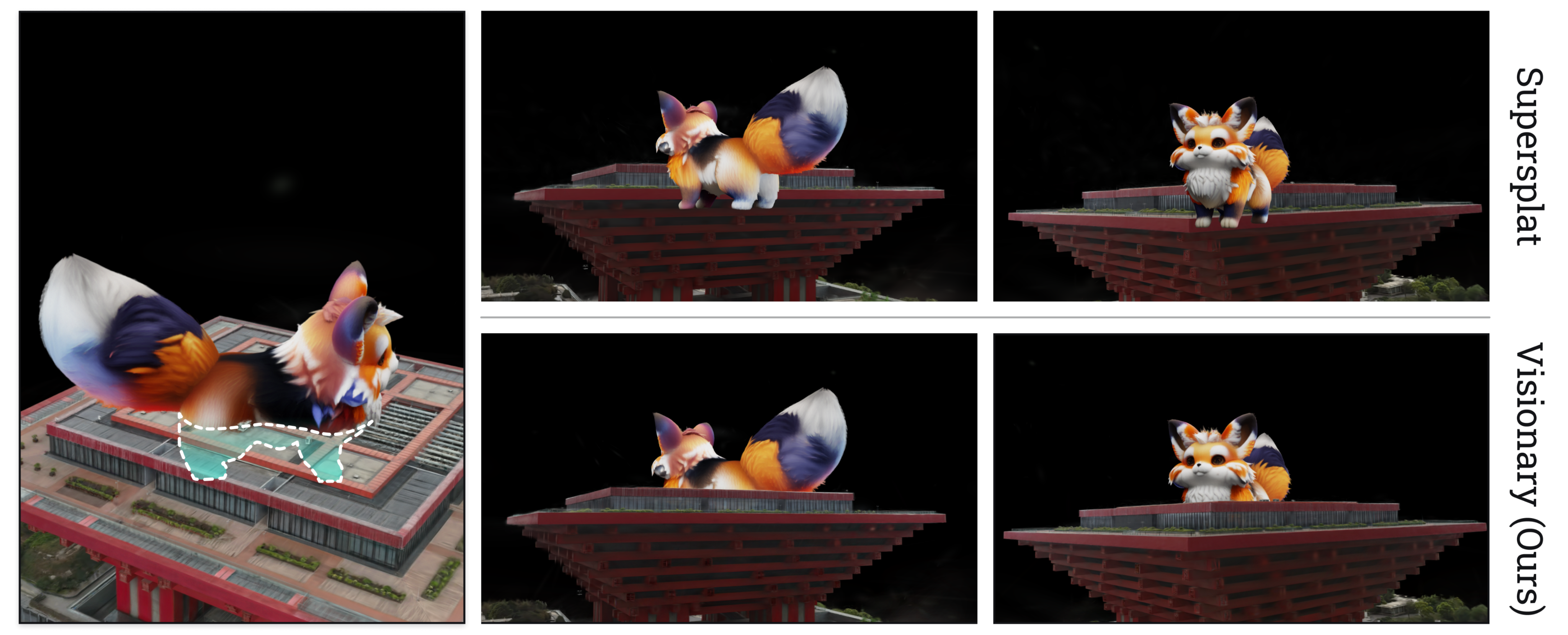}
  \caption{\textbf{Wrong visualization caused by local sorting in \texttt{Supersplat}.}
  Without a global ordering, overlapping Gaussians across partitions may be composited in an incorrect order. Visionary's efficient global sorting avoids this issue. The version of \texttt{Supersplat} tested here is the latest available, \texttt{v2.15.1}.}
  \label{fig:supersplat_local_sort_artifacts}
\end{figure}

\subsection{Robustness under Rapid Viewpoint Changes}
\label{sec:exp_lazy_sort}
\paragraph{\texttt{SparkJS} lazy sorting.} Besides average frame time, interactive viewers must remain visually stable under fast user input.
\texttt{SparkJS} adopts a \emph{lazy sorting} strategy to reduce CPU overhead by reusing a stale ordering and updating it incrementally, which means instead of re-sorting splats every frame in lockstep with rendering, it performs asynchronous updates that are amortized over multiple frames, so the ordering can lag behind the current view. However, when the camera rotates quickly, the depth ordering can change drastically between frames, and the approximate ordering may become invalid. This leads to incorrect alpha compositing and visible artifacts (\eg, popping, streaking, and inconsistent transparency), as shown in~\Cref{fig:spark_lazy_sort_artifacts}.

Visionary avoids this kind of failure by performing a true \emph{global per-frame sorting} on the GPU for all visible splats.
Although sorting is non-trivial, our \texttt{WebGPU} compute implementation keeps the overhead small and stable,
and preserves correct back-to-front compositing under fast camera motion.

\subsection{Composition Correctness vs.\ \texttt{Supersplat}}
\label{sec:exp_supersplat}
\paragraph{Why global sorting matters?}
\texttt{SuperSplat} improves performance by avoiding a full global ordering and instead relies on \emph{local sorting}
(\eg, sorting within partitions). While efficient, such a strategy is not equivalent to a single global back-to-front ordering. When Gaussians from different partitions overlap substantially, the lack of a global order leads to inconsistent blending between partitions, yielding depth-inconsistent transparency (\Cref{fig:supersplat_local_sort_artifacts}).

Visionary uses a unified global buffer and performs one global sort for \emph{all} valid splats across all loaded models
within the same frame, ensuring correct compositing even when multiple 3DGS assets (or 3DGS + mesh) are rendered together.

\subsection{Overhead of ONNX-based Gaussian pre-decoding}
\label{sec:exp_onnx_overhead}
Finally, we report the runtime feasibility of integrating representative dynamic or structured 3DGS variants through
our \texttt{ONNX}-based \texttt{Gaussian Generator} contract.
For MLP-based 3DGS (\eg, Scaffold-GS~\cite{lu2024scaffold}) and deformable 3DGS or 4DGS~\cite{wu20244d}, the per-frame decoding runs in real-time across different scenes as in \Cref{tab:MLP_time}.
For animatable avatars~\cite{hu2024gauhuman,qiu2025lhm,zhan2025r3}, we observe per-frame inference in the range of 7--8\si{ms} depending on the specific model as in \Cref{tab:Avater_time}. Note that avatars typically require fewer Gaussians.
These results indicate that a single browser-resident pipeline can support both fast rendering and per-frame neural updates.

For 3DGS variants beyond the original 3DGS implementation and diffusion-based post-processing effects, there are currently no other web viewers that support comparable functionality; we therefore refer readers to the accompanying video demos and our online editor for qualitative inspection and interactive exploration.

\begin{table}[!t]
\centering
\setlength{\tabcolsep}{9pt}
\caption{\textbf{Inference time of MLP-based 3DGS and 4DGS.}}
\label{tab:MLP_time}
\begin{tabular}{lcccc}
\toprule
& \multicolumn{2}{c}{Scaffold-GS~\cite{lu2024scaffold}} & \multicolumn{2}{c}{4DGS~\cite{wu20244d}} \\
\cmidrule(lr){2-3}
\cmidrule(lr){4-5}
\# GS [M] & 2.49 & 4.56 & 0.03 & 0.06\\
Time [\si{ms}] & 9.29 &16.10 & 4.76 & 7.93\\
\bottomrule
\end{tabular}
\end{table}

\begin{table}[!t]
\centering
\setlength{\tabcolsep}{9pt}
\caption{\textbf{Inference time of Avatar.}}
\label{tab:Avater_time}
\begin{tabular}{lcccccc}
\toprule
& \multicolumn{3}{c}{Gauhuman~\cite{hu2024gauhuman}} & \multicolumn{3}{c}{R$^3$-Avatar~\cite{zhan2025r3}} \\
\cmidrule(lr){2-4}
\cmidrule(lr){5-7}
\# Instances & 1 & 5 & 10 & 1 & 5 & 10 \\
\# GS [M] & 0.04 &0.20 &0.40 & 0.02 & 0.12 & 0.23 \\
Time [\si{ms}] & 7.97 & 28.60 &55.63 & 7.46 & 27.10 & 53.35 \\
\bottomrule
\end{tabular}
\end{table}



\section{Discussion}
\subsection{Large World Model \& Future}
The definition and implementation of world models are still under active debate, with no single absolutely dominant paradigm. Some approaches, such as \href{https://www.worldlabs.ai/blog/marble-world-model}{Marble} and FlashWorld~\cite{li2025flashworld}, generate 3DGS scenes from images or text and emphasize reconstructive, geometrically consistent representations. Others, such as \href{https://deepmind.google/blog/genie-3-a-new-frontier-for-world-models/}{Genie 3} and \href{https://www.worldlabs.ai/blog/rtfm}{RTFM}, follow a video-generation paradigm that prioritizes creative synthesis and open-ended dynamics. We believe that the next generation of world model renderers should combine the strengths of both paradigms, forming a closed loop between generation and reconstruction. Visionary is designed as an initial step toward this unified framework.

Building on the current architecture, several promising directions can be explored. First, we plan to improve physical interaction by integrating collision detection and further coupling with mesh-based pipelines~\cite{guo2025tagsplat}. Second, we aim to incorporate physics-aware modeling, in which 3DGS representations are combined with methods such as the Material Point Method (MPM) to simulate realistic dynamics~\cite{xie2024physgaussian,lin2025omniphysgs,liu2024dynamic}. Third, we will investigate spatially grounded 3D agents built on top of multimodal language models~\cite{yang2025thinking,brown2025sims,yang2025cambrian}, enabling reasoning and interaction within complex environments. Finally, Visionary can serve as a bridge to downstream applications by interfacing with vectorized physics simulators, such as Isaac Gym~\cite{makoviychuk2021isaac}, and by integrating relighting~\cite{gao2024relightable,liang2024gs} and domain adaptation techniques to support Sim-to-Real transfer for embodied AI~\cite{escontrela2025gaussgym,jia2025discoverse}.

\subsection{Limitations}
\texttt{WebGPU} and \texttt{ONNX} runtimes are still evolving, which may lead to compatibility and stability differences across browsers and operating systems. In addition, current browser security policies impose CPU memory constraints, limiting the size of models that can be executed fully in-browser using off-the-shelf tools. As a result, while small-scale to medium-scale networks can be integrated into the pipeline,  some post-processing steps currently remain offline.

\section{Conclusion}
Visionary addresses several fundamental barriers in neural rendering platforms, including the lack of effective web-native computation, limited extensibility, and heavy system dependencies. By combining a modern, compute-capable graphics API (\texttt{WebGPU}) with a unified \texttt{ONNX}-based integration contract, Visionary enables real-time, in-browser rendering of diverse 3DGS variants in a portable, lightweight, and research-friendly manner, with support for generative post-processing. Beyond serving as a renderer, it acts as a unified carrier for both reconstructive and generative world modeling paradigms, establishing an open and extensible foundation for future advances in spatial intelligence, embodied agents, and interactive 3D environments.
\section*{Acknowledgments}
This work was partially supported by the Shanghai AI Laboratory. We thank Kang An$^{2}$ and Tianchen Hao$^{2}$ for their contributions to the project during their internships at the Shanghai AI Laboratory.




\bibliographystyle{abbrv}
\bibliography{refs}

\end{document}